\documentclass[11pt,a4paper]{article}
\usepackage{paralist}

\usepackage{authblk}
\usepackage[hyperref]{emnlp2018}
\usepackage{times}
\usepackage{latexsym}

\usepackage{url}

\aclfinalcopy 

\usepackage{amsfonts}
\usepackage{amsmath}
\usepackage{color}
\usepackage{graphicx}
\usepackage{xspace}
\usepackage{amsfonts}
\usepackage{algorithm}
\usepackage[noend]{algorithmic}
\usepackage{graphicx}
\usepackage{listings}
\usepackage{verbatim}
\usepackage{varwidth}
\def\xorig{$\mathbf{x}_{orig}$\xspace}
\def\xadv{$\mathbf{x}_{adv}$\xspace}
\def\xcur{$\mathbf{x}_{cur}$\xspace}

\def\xadv{\mathbf{x}_{adv} }
\def\xorig{\mathbf{x}_{orig}}
\newcommand{\orw}[1]{\textcolor{green}{\textbf{#1}}}
\newcommand{\mow}[1]{\textcolor{red}{\textbf{#1}}}

\DeclareMathOperator*{\argmax}{arg\,max}

\def\perturb{\texttt{Perturb}\xspace}

\title{Generating Natural Language Adversarial Examples}
\author{ \vspace{-0.8cm} \\ Moustafa Alzantot\textsuperscript{1}\thanks{Moustafa Alzantot and Yash Sharma contribute equally to this work.} , Yash Sharma\textsuperscript{2$*$},  Ahmed Elgohary\textsuperscript{3},\\ Bo-Jhang Ho\textsuperscript{1}, Mani B. Srivastava\textsuperscript{1}, Kai-Wei Chang\textsuperscript{1}\\ \\
\textsuperscript{1}Department of Computer Science, University of California, Los Angeles (UCLA) \\
\texttt{\{malzantot, bojhang, mbs, kwchang\}@ucla.edu}\\
\textsuperscript{2}Cooper Union \texttt{sharma2@cooper.edu} \\
\textsuperscript{3}Computer Science Department, University of Maryland \texttt{elgohary@cs.umd.edu} 
}

\begin{document}
 \maketitle
\begin{abstract}
Deep neural networks (DNNs) are vulnerable to adversarial examples, perturbations to correctly classified examples which can cause the model to misclassify. In the image domain, these perturbations are often virtually indistinguishable to human perception, causing humans and state-of-the-art models to disagree. However, in the natural language domain, small perturbations are clearly perceptible, and the replacement of a single word can drastically alter the semantics of the document. Given these challenges, we use a black-box population-based optimization algorithm to generate semantically and syntactically similar adversarial examples that fool well-trained sentiment analysis and textual entailment models with success rates of 97\% and 70\%, respectively. We additionally demonstrate that 92.3\% of the successful sentiment analysis adversarial examples are classified to their original label by 20 human annotators, and that the examples are perceptibly quite similar. Finally, we discuss an attempt to use adversarial training as a defense, but fail to yield improvement, demonstrating the strength and diversity of our adversarial examples. We hope our findings encourage researchers to pursue improving the robustness of DNNs in the natural language domain.  
\end{abstract}

\section{Introduction}
\vspace{-0.1cm}
Recent research has found that deep neural networks (DNNs) are vulnerable to \textit{adversarial examples}~\cite{goodfellow2014explaining, szegedy2013intriguing}. The existence of adversarial examples has been shown in image classification~\cite{szegedy2013intriguing} and speech recognition~\cite{carlini2018audio}. In this work, we demonstrate that adversarial examples can be constructed in the context of natural language. Using a black-box population-based optimization algorithm, we successfully generate both semantically and syntactically similar adversarial examples against models trained on both the IMDB~\cite{maas-EtAl:2011:ACL-HLT2011} sentiment analysis task and the Stanford Natural Language Inference (SNLI)~\cite{bowman2015large} textual entailment task. In addition, we validate that the examples are both correctly classified by human evaluators and similar to the original via a human study. Finally, we attempt to defend against said adversarial attack using adversarial training, but fail to yield any robustness, demonstrating the strength and diversity of the generated adversarial examples. 

Our results show that by minimizing the semantic and syntactic dissimilarity, an attacker can perturb examples such that humans correctly classify, but high-performing models misclassify. We are open-sourcing our attack\footnote{\url{https://github.com/nesl/nlp\_adversarial\_examples}} to encourage research in training DNNs robust to adversarial attacks in the natural language domain.

\section{Natural Language Adversarial Examples}
\vspace{-0.1cm}
Adversarial examples have been explored primarily in the image recognition domain. Examples have been generated through solving an optimization problem, attempting to induce misclassification while minimizing the perceptual distortion~\cite{szegedy2013intriguing,carlini2017towards,ead,ead_madry}. Due to the computational cost of such approaches, fast methods were introduced which, either in one-step or iteratively, shift all pixels simultaneously until a distortion constraint is reached~\cite{goodfellow2014explaining,kurakin2016adversarial_ICLR,madry2017towards}. Nearly all popular methods are gradient-based.

Such methods, however, rely on the fact that adding small perturbations to many pixels in the image will not have a noticeable effect on a human viewer. This approach obviously does not transfer to the natural language domain, as all changes are perceptible. Furthermore, unlike continuous image pixel values, words in a sentence are discrete tokens. Therefore, it is not possible to compute the gradient of the network loss function with respect to the input words. A straightforward workaround is to project input sentences into a continuous space (e.g. word embeddings) and consider this as the model input. However, this approach also fails because it still assumes that replacing \textit{every} word with words nearby in the embedding space will not be noticeable. Replacing words without accounting for syntactic coherence will certainly lead to improperly constructed sentences which will look odd to the reader.

Relative to the image domain, little work has been pursued for generating natural language adversarial examples. Given the difficulty in generating semantics-preserving perturbations, distracting sentences have been added to the input document in order to induce misclassification~\cite{jia2017adversarial}. In our work, we attempt to generate semantically and syntactically similar adversarial examples, via word replacements, resolving the aforementioned issues. Minimizing the number of word replacements necessary to induce misclassification has been studied in previous work~\cite{papernot}, however without consideration given to semantics or syntactics, yielding incoherent generated examples.
In recent work, there have been a few attempts at generating adversarial examples for language tasks by using back-translation~\cite{iyyer-2018-controlled}, exploiting machine-generated rules~\cite{ribeiro2018semantically}, and searching in underlying semantic space~\cite{zhao2017generating}. In addition, while preparing our submission, we became aware of recent work which target a similar contribution~\cite{iclr,hotflip}. We treat these contributions as parallel work.
\vspace{-0.2\baselineskip}
\section{Attack Design}
\vspace{-0.2\baselineskip}
\subsection{Threat model}
\vspace{-0.1cm}
We assume the attacker has black-box access to the target model; the attacker is not aware of the model architecture, parameters, or training data, and is only capable of querying the target model with supplied inputs and obtaining the output predictions and their confidence scores. This setting has been extensively studied in the image domain~\cite{papernot_blackbox, zoo, genattack}, but has yet to be explored in the context of natural language.

%
%

\subsection{Algorithm}
\vspace{-0.1cm}
To avoid the limitations of gradient-based attack methods, we design an algorithm for constructing adversarial examples with the following goals in mind. We aim to minimize the number of modified words between the original and adversarial examples, but only perform modifications which retain semantic similarity with the original and syntactic coherence. To achieve these goals, instead of relying on gradient-based optimization, 
we developed an attack algorithm that exploits population-based gradient-free optimization via genetic algorithms. 

An added benefit of using gradient-free optimization is enabling use in the black-box case; gradient-reliant algorithms are inapplicable in this case, as they are dependent on the model being differentiable and the internals being accessible~\cite{papernot,hotflip}. 

Genetic algorithms are inspired by the process of natural selection, iteratively evolving a population of candidate solutions towards better solutions. The population of each iteration is a called a \textit{generation}. In each generation, the quality of population members is evaluated using a \textit{fitness} function. ``Fitter'' solutions are more likely to be selected for breeding the next generation. The next generation is generated through a combination of \textit{crossover} and \textit{mutation}. Crossover is the process of taking more than one parent solution and producing a child solution from them; it is analogous to reproduction and biological crossover. Mutation is done in order to increase the diversity of population members and provide better exploration of the search space. Genetic algorithms are known to perform well in solving combinatorial optimization problems~\cite{anderson1994genetic, muhlenbein1989parallel}, and due to employing a population of candidate solutions, these algorithms can find successful adversarial examples with fewer modifications.

\paragraph*{\perturb Subroutine:} In order to explain our algorithm, we first introduce the subroutine \perturb. This subroutine accepts an input sentence \xcur which can be either a modified sentence or the same as $\xorig$. It randomly selects a word $w$ in the sentence \xcur and then selects a suitable replacement word that has similar semantic meaning, fits within the surrounding context, and increases the $target$ label prediction score. \\In order to select the best replacement word, \perturb  applies the following steps:
\begin{compactitem}
\item Computes the $N$ nearest neighbors of the selected word according to the distance in the GloVe embedding space~\cite{pennington2014glove}. We used euclidean distance, as we did not see noticeable improvement using cosine. We filter out candidates with distance to the selected word greater than $\delta$. We use the counter-fitting method presented in~\cite{mrkvsic2016counter} to post-process the adversary's GloVe vectors to ensure that the nearest neighbors are synonyms. The resulting embedding is independent of the embeddings used by victim models.
\item Second, we use the Google 1 billion words language model ~\cite{chelba2013one} to filter out words that do not fit within the context surrounding the word $w$ in \xcur. We do so by ranking the candidate words based on their language model scores when fit within the replacement context, and keeping only the top $K$ words with the highest scores.
\item From the remaining set of words, we pick the one that will maximize the target label prediction probability when it replaces the word $w$ in \xcur. \item Finally, the selected word is inserted in place of $w$, and \perturb returns the resulting sentence.
\end{compactitem} 

The selection of which word to replace in the input sentence is done by random sampling with probabilities proportional to the number of neighbors each word has within Euclidean distance $\delta$ in the counter-fitted embedding space, encouraging the solution set to be large enough for the algorithm to make appropriate modifications. We exclude common articles and prepositions (e.g. a, to) from being selected for replacement.

\begin{algorithm}[htb]
\setlength\belowcaptionskip{-10pt} 
   \caption{Finding adversarial examples}
   \label{alg:main}
\begin{algorithmic}
 \FOR{$i=1,...,S$ in population}
   \STATE $\mathcal{P}^{0}_i \leftarrow \perturb(\xorig, target)  $
   \ENDFOR
	\FOR{$g=1,2...G \text{ generations}$}
   	       \FOR{$i=1,...,S$ in population}
   	     \STATE $F^{g-1}_i = f(\mathcal{P}^{g-1}_i)_{target}$
   	     \ENDFOR
   	     \STATE $\xadv = \mathcal{P}^{g-1}_{\argmax_{j} F^{g-1}_j}$
   	     \IF {$\argmax_c f(\xadv)_c == t$ }
   	     \STATE {\bfseries return} $\xadv$ $\triangleright$  \COMMENT{Found successful attack}
   	     \ELSE
	\STATE $\mathcal{P}^{g}_1 = \{ \xadv \}$ 
	\STATE $p = Normalize(F^{g-1})$ \FOR{$i=2,...,S$ in population}
\STATE $\text{Sample } parent_1 \text{ from }  \mathcal{P}^{g-1} \text{ with probs } p$\STATE $\text{Sample } parent_2 \text{ from }  \mathcal{P}^{g-1} \text{  with probs } p$\STATE$child$ = $Crossover$($parent_1$, $parent_2$)\STATE$child_{mut} = \perturb(child, target)$
\STATE $\mathcal{P}^{g}_i  = \{ child_{mut} \}$
\ENDFOR
\ENDIF
\ENDFOR
\end{algorithmic}
\end{algorithm}

\paragraph*{Optimization Procedure:} The optimization algorithm can be seen in Algorithm~\ref{alg:main}. The algorithm starts by creating the initial generation $\mathcal{P}^{0}$ of size $S$  by calling the \perturb subroutine $S$ times to create a set of distinct modifications to the original sentence. Then, the fitness of each population member in the current generation is computed as the target label prediction probability, found by querying the victim model function $f$. If a population member's predicted label is equal to the target label, the optimization is complete. Otherwise, pairs of population members from the current generation are randomly sampled with probability proportional to their fitness values. A new \textit{child} sentence is then synthesized from a pair of parent sentences by independently sampling from the two using a uniform distribution. Finally, the \perturb subroutine is applied to the resulting children.

\section{Experiments}
\vspace{-0.15cm}
%
%
\begin{table*}[!t]
\centering

\begin{tabular}{|p{16cm}|}
\hline
Original Text Prediction = \textbf{Negative}. (Confidence = 78.0\%) \\
\hline 
\textit{This movie had \orw{terrible} acting, \orw{terrible} plot, and \orw{terrible} choice of actors. (Leslie Nielsen ...come on!!!) the one part I \orw{considered} slightly funny was the battling FBI/CIA agents, but because the audience was mainly \orw{kids} they didn't understand that theme.} \\
\hline
\hline
Adversarial Text Prediction = \textbf{Positive}. (Confidence = 59.8\%)  \\
\hline
\textit{This movie had \mow{horrific} acting, \mow{horrific} plot, and \mow{horrifying} choice of actors. (Leslie Nielsen ...come on!!!) the one part I \mow{regarded} slightly funny was the battling FBI/CIA agents, but because the audience was mainly \mow{youngsters} they didn't understand that theme.} \\
\hline
\end{tabular}
\vspace{-4mm}
\caption{Example of attack results for the sentiment analysis task. Modified words are highlighted in green and red for the original and adversarial texts, respectively.}
\label{tab:output_sample1}
\end{table*}

\begin{table*}[!t]
\setlength\belowcaptionskip{-5pt}
\centering
\begin{tabular}{|p{16cm}|}
\hline
Original Text Prediction: {\bf Entailment} (Confidence = 86\%) \\
\hline
\textbf{Premise:} \textit{A runner wearing purple strives for the finish line.}  \\
\textbf{Hypothesis:} \textit{A \orw{runner} wants to head for the finish line.} \\
\hline
\hline
Adversarial Text Prediction: {\bf Contradiction} (Confidence = 43\%) \\
\hline
\textbf{Premise:} \textit{A runner wearing purple strives for the finish line.}  \\
\textbf{Hypothesis:}  \textit{ A \mow{racer} wants to head for the finish line.} \\
\hline
\end{tabular}
\caption{Example of attack results for the textual entailment task. Modified words are highlighted in green and red for the original and adversarial texts, respectively.}
\label{tab:output_sample2}
\end{table*}


\begin{table*}[!t]
\setlength\belowcaptionskip{-10pt}
\centering
\begin{tabular}{c|c|c||c|c|}
\cline{2-5}
 & \multicolumn{2}{c||}{Sentiment Analysis}& \multicolumn{2}{|c|}{Textual Entailment}\\
 & \% success & \% modified & \
\% success & \% modified \\
  \hline
 \multicolumn{1}{|c|}{\perturb baseline} & 52\% & 19\% & -- & -- \\
 \hline
 \multicolumn{1}{|c|}{Genetic attack} &  97\% & 14.7\% & 70\% & 23\% \\
 \hline
\end{tabular}
\caption{Comparison between the attack success rate and mean percentage of modifications required by the genetic attack and perturb baseline for the two tasks.}
\label{tab:asr}
\end{table*}

To evaluate our attack method, we trained models for the sentiment analysis and textual entailment classification tasks. For both models, each word in the input sentence is first projected into a fixed 300-dimensional vector space using GloVe~\cite{pennington2014glove}. Each of the models used are based on popular open-source benchmarks, and can be found in the following repositories\footnote{\url{https://github.com/keras-team/keras/blob/master/examples/imdb_lstm.py}}\footnote{\url{https://github.com/Smerity/keras_snli/blob/master/snli_rnn.py}}. Model descriptions are given below. 

\textbf{Sentiment Analysis:} We trained a sentiment analysis model using the IMDB dataset of movie reviews~\cite{maas-EtAl:2011:ACL-HLT2011}. The IMDB dataset consists of 25,000 training examples and 25,000 test examples. The LSTM model is composed of 128 units, and the outputs across all time steps are averaged and fed to the output layer. The test accuracy of the model is 90\%, which is relatively close to the state-of-the-art results on this dataset. 

\textbf{Textual Entailment:} We trained a textual entailment model using the Stanford Natural Language Inference (SNLI) corpus~\cite{bowman2015large}. The model passes the input through a ReLU ``translation'' layer~\cite{bowman2015large}, which encodes the premise and hypothesis sentences by performing a summation over the word embeddings, concatenates the two sentence embeddings, and finally passes the output through 3 600-dimensional ReLU layers before feeding it to a 3-way softmax. The model predicts whether the premise sentence entails, contradicts or is neutral to the hypothesis sentence. The test accuracy of the model is~83\% which is also relatively close to the state-of-the-art~\cite{chen2017enhanced}. 
\vspace{-0.25cm}
\subsection{Attack Evaluation Results}
\label{sec:evaluation}
\vspace{-0.15cm}
We randomly sampled 1000, and 500 \textit{correctly classified} examples from the test sets of the two tasks to evaluate our algorithm. Correctly classified examples were chosen to limit the accuracy levels of the victim models from confounding our results. For the sentiment analysis task, the attacker aims to divert the prediction result from positive to negative, and vice versa. For the textual entailment task, the attacker is only allowed to modify the hypothesis, and aims to divert the prediction result from `entailment' to `contradiction', and vice versa. We limit the attacker to maximum $G=20$ iterations, and fix the hyperparameter values to $S=60$, $N=8$, $K=4$, and $\delta=0.5$. We also fixed the maximum percentage of allowed changes to the document to be 20\% and 25\% for the two tasks, respectively. If increased, the success rate would increase but the mean quality would decrease. If the attack does not succeed within the iterations limit or exceeds the specified threshold, it is counted as a failure. 

Sample outputs produced by our attack are shown in tables~\ref{tab:output_sample1} and~\ref{tab:output_sample2}. Additional outputs can be found in the supplementary material. Table~\ref{tab:asr} shows the attack success rate and mean percentage of modified words on each task. We compare to the \perturb baseline, which greedily applies the \perturb subroutine, to validate the use of population-based optimization. As can be seen from our results, we are able to achieve high success rate with a limited number of modifications on both tasks. In addition, the genetic algorithm significantly outperformed the \perturb baseline in both success rate and percentage of words modified, demonstrating the additional benefit yielded by using population-based optimization. Testing using a single TitanX GPU, for sentiment analysis and textual entailment, we measured average runtimes on success to be 43.5 and 5 seconds per example, respectively. The high success rate and reasonable runtimes demonstrate the practicality of our approach, even when scaling to long sentences, such as those found in the IMDB dataset.  

Speaking of which, our success rate on textual entailment is lower due to the large disparity in sentence length. On average, hypothesis sentences in the SNLI corpus are 9 words long, which is very short compared to IMDB (229 words, limited to 100 for experiments). With sentences that short, applying successful perturbations becomes much harder, however we were still able to achieve a success rate of 70\%. For the same reason, we didn't apply the \perturb baseline on the textual entailment task, as the \perturb baseline fails to achieve any success under the limits of the maximum allowed changes constraint. 


\vspace{-0.15cm}
\subsection{User study}
\vspace{-0.15cm}
We performed a user study on the sentiment analysis task with 20 volunteers to evaluate how perceptible our adversarial perturbations are. Note that the number of participating volunteers is significantly larger than used in previous studies~\cite{jia2017adversarial, hotflip}. The user study was composed of two parts. First, we presented 100 adversarial examples to the participants and asked them to label the sentiment of the text (i.e., positive or negative.) 92.3\% of the responses matched the original text sentiment, indicating that our modification did not significantly affect human judgment on the text sentiment. Second, we prepared 100 questions, each question includes the original example and the corresponding adversarial example in a pair. Participants were asked to judge the similarity of each pair on a scale from 1 (very similar) to 4 (very different). The average rating is $2.23\pm0.25$, which shows the perceived difference is also small.
\vspace{-0.20cm}
\subsection{Adversarial Training}
\vspace{-0.20cm}
\label{sec:defense}
The results demonstrated in section~\ref{sec:evaluation} raise the following question: How can we defend against these attacks? We performed a preliminary experiment to see if adversarial training~\cite{madry2017towards}, the only effective defense in the image domain, can be used to lower the attack success rate. We generated 1000 adversarial examples on the cleanly trained sentiment analysis model using the IMDB training set, appended them to the existing training set, and used the updated dataset to adversarially train a model from scratch. We found that adversarial training provided no additional robustness benefit in our experiments using the test set, despite the fact that the model achieves near 100\% accuracy classifying adversarial examples included in the training set. These results demonstrate the diversity in the perturbations generated by our attack algorithm, and illustrates the difficulty in defending against adversarial attacks. We hope these results inspire further work in increasing the robustness of natural language models.
\vspace{-0.22cm}
\section{Conclusion}
\vspace{-0.22cm}
We demonstrate that despite the difficulties in generating imperceptible adversarial examples in the natural language domain, semantically and syntactically similar adversarial examples can be crafted using a black-box population-based optimization algorithm, yielding success on both the sentiment analysis and textual entailment tasks. Our human study validated that the generated examples were indeed adversarial and perceptibly quite similar. We hope our work encourages researchers to pursue improving the robustness of DNNs in the natural language domain.
\vspace{-0.25cm}
\section*{Acknowledgement}
\vspace{-0.25cm}
This research was supported in part by  the U.S. Army Research Laboratory and the UK Ministry of Defence under Agreement Number W911NF-16-3-0001, the National Science Foundation under award \# CNS-1705135, OAC-1640813, and IIS-1760523, and the NIH Center of Excellence for Mobile Sensor Data-to-Knowledge (MD2K) under award 1-U54EB020404-01. Ahmed Elgohary is funded by an IBM PhD Fellowship. Any findings in this material are those of the author(s) and do not reflect the views of any of the above funding agencies. The U.S. and U.K. Governments are authorized to reproduce and distribute reprints for Government purposes notwithstanding any copyright notation hereon.

\bibliography{emnlp2018}
\bibliographystyle{acl_natbib_nourl}
\clearpage
\onecolumn
\textbf{\Large Supplemental Materials: Generating Natural Language Adversarial Examples}\\
\section*{Additional Sentiment Analysis Results}

Table~\ref{tab:output_sample1} shows an additional set of attack results against the sentiment analysis model described in our paper.

\begin{table*}[!h]
\centering
\begin{tabular}{|p{16cm}|}
\hline
Original Text Prediction = \textbf{Positive}. (Confidence = 78\%) \\
\hline The promise of Martin Donovan playing Jesus was, \orw{quite} \orw{honestly}, enough to get me to see the film. Definitely worthwhile; clever and funny without overdoing it. The low quality filming was \orw{probably} an appropriate effect but ended up being a little too jarring, and the ending sounded more like a PBS program than Hartley. Still, too many memorable lines and great moments for me to judge it harshly. \\
\hline
Adversarial Text Prediction = \textbf{Negative}. (Confidence = 59.9\%)  \\
\hline
The promise of Martin Donovan playing Jesus was, \mow{utterly} \mow{frankly}, enough to get me to see the film. Definitely worthwhile; clever and funny without overdoing it. The low quality filming was \mow{presumably} an appropriate effect but ended up being a little too jarring, and the ending sounded more like a PBS program than Hartley. Still, too many memorable lines and great moments for me to judge it harshly. \\
\hline
\end{tabular}

\vspace{0.3cm}

\begin{tabular}{|p{16cm}|}
\hline
Original Text Prediction = \textbf{Negative}. (Confidence = 74.30\%) \\
\hline 
Some \orw{sort} of accolade must be given to  `Hellraiser: Bloodline'. It's \orw{actually} out full-mooned Full Moon. It bears all the marks of, say, your `demonic toys' or `puppet master' series, without their \orw{dopey}, uh, charm? Full Moon can get away with \orw{silly} product because they know it's silly. These Hellraiser things, man, do they ever take themselves seriously. This \orw{increasingly} \orw{stupid} franchise (\orw{though} not nearly as \orw{stupid} as I am for having \orw{watched} it) once made up for its low budgets by being stylish. Now it's just ish. \\
\hline
Adversarial Text Prediction = \textbf{Positive}. (Confidence = 51.03\%)  \\
\hline
Some \mow{kind} of accolade must be given to  `Hellraiser: Bloodline'. it's \mow{truly} out full-mooned Full Moon. It bears all the marks of, say, your `demonic toys' or `puppet master' series, without their \mow{silly}, uh, charm? Full Moon can get away with \mow{daft} product because they know it's silly. These Hellraiser things, man, do they ever take themselves seriously. This \mow{steadily} \mow{daft} franchise (\mow{whilst} not nearly as \mow{daft} as i am for having \mow{witnessed} it) once made up for its low budgets by being stylish. Now it's just ish. \\
\hline
\end{tabular}

\vspace{0.3cm}
\begin{tabular}{|p{16cm}|}
\hline
Original Text Prediction = \textbf{Negative}. (Confidence = 50.53\%) \\
\hline 
Thinly-cloaked retelling of the garden-of-eden story -- nothing new, nothing shocking, although I feel that is what the filmmakers were going for. The idea is \orw{trite}. Strong performance from Daisy Eagan, that's about it. I believed she was 13, and I was interested in her character, the rest left me cold. \\
\hline
Adversarial Text Prediction = \textbf{Positive}. (Confidence = 63.04\%)  \\
\hline
Thinly-cloaked retelling of the garden-of-eden story -- nothing new, nothing shocking, although I feel that is what the filmmakers were going for. The idea is \orw{petty}. Strong performance from Daisy Eagan, that's about it. I believed she was 13, and I was interested in her character, the rest left me cold. \\
\hline
\end{tabular}

\caption{Example of attack results against the sentiment analysis model. Modified words are highlighted in green and red for the original and adversarial texts, respectively.}
\label{tab:output_sample1}
\end{table*}

\clearpage
\section*{Additional Textual Entailment Results}

Table~\ref{tab:output_sample2} shows an additional set of attack results against the textual entailment model described in our paper.
\begin{table*}[!h]
\centering
\begin{tabular}{|p{16cm}|}
\hline
Original Text Prediction: \textbf{Contradiction} (Confidence = 91\%) \\
\hline
\textbf{Premise:} A man and a woman stand in front of a Christmas tree contemplating a single thought.   \\
\textbf{Hypothesis:} Two \orw{people} \orw{talk} loudly in front of a cactus.\\
\hline
\hline
Adversarial Text Prediction: \textbf{Entailment} (Confidence = 51\%) \\
\hline
\textbf{Premise:} A man and a woman stand in front of a Christmas tree contemplating a single thought.  \\
\textbf{Hypothesis:}  Two \mow{humans} \mow{chitchat} loudly in front of a cactus.
\\
\hline
\end{tabular}
\vspace{0.3cm}

\begin{tabular}{|p{16cm}|}
\hline
Original Text Prediction: {\bf Contradiction} (Confidence = 94\%) \\
\hline
\textbf{Premise:} A young girl wearing yellow shorts and a white tank top using a cane pole to fish at a small pond.  \\
\textbf{Hypothesis:} A girl wearing a \orw{dress} looks off a \orw{cliff}.\\
\hline
\hline
Adversarial Text Prediction: {\bf Entailment} (Confidence = 40\%) \\
\hline
\textbf{Premise:} A young girl wearing yellow shorts and a white tank top using a cane pole to fish at a small pond.  \\
\textbf{Hypothesis:}  A girl wearing a \mow{skirt} looks off a \mow{ravine}.\\
\hline
\end{tabular}
\vspace{0.3cm}

\begin{tabular}{|p{16cm}|}
\hline
Original Text Prediction: \textbf{Entailment} (Confidence = 86\%) \\
\hline
\textbf{Premise:} A large group of protesters are walking down the street with signs.  \\
\textbf{Hypothesis:} Some people are holding up \orw{signs} of protest in the street.\\
\hline
\hline
Adversarial Text Prediction: {\bf Contradiction} (Confidence = 43\%) \\
\hline
\textbf{Premise:} A large group of protesters are walking down the street with signs.  \\
\textbf{Hypothesis:}   Some people are holding up \mow{signals} of protest in the street.\\
\hline
\end{tabular}

\caption{Example of attack results against the textual entailment model. Modified words are highlighted in green and red for the original and adversarial texts, respectively.}
\label{tab:output_sample2}
\end{table*}

\end{document}